\pdfoutput=1

\documentclass[11pt]{article}

\usepackage{ACL2023}

\usepackage{times}
\usepackage{latexsym}
\usepackage{multicol}
\usepackage{multirow}
\usepackage{subfigure}

\usepackage[T1]{fontenc}

\usepackage[utf8]{inputenc}

\usepackage{microtype}

\usepackage{inconsolata}
\usepackage{booktabs}
\usepackage{multirow}
\usepackage{graphicx}
\usepackage{amsmath}
\usepackage{amsfonts}
\usepackage{algpseudocode}
\usepackage{algorithmicx}
\usepackage{algorithm}
\definecolor{ForestGreen}{HTML}{009B55}
\usepackage{pifont}
\newcommand{\xmark}{\ding{55}}%

\title{Speech-Text Dialog Pre-training for Spoken Dialog Understanding \\ with Explicit Cross-Modal Alignment}

\author{
Tianshu Yu$^{123*}$, Haoyu Gao$^{134}$\thanks{\ \ Equal contribution. This work was conducted when Tianshu Yu and Haoyu Gao were interning at Alibaba.}, Ting-En Lin$^{3}$, Min Yang$^{1\dagger}$, \\  
\textbf{Yuchuan Wu$^{3}$, Wentao Ma$^{3}$, Chao Wang$^{3}$, Fei Huang$^{3}$, Yongbin Li$^{3}$\thanks{\textsuperscript{\textdagger} Min Yang and Yongbin Li are corresponding authors.}}  \\
  $^{1}$Shenzhen Institute of Advanced Technology, Chinese Academy of Sciences \\ 
  $^{2}$University of Chinese Academy of Sciences\\
  $^{3}$Alibaba Group \quad
  $^{4}$University of Science and Technology of China \\
  \texttt{\{ts.yu,min.yang\}@siat.ac.cn} \\ 
  \texttt{\{ghy385779,ting-en.lte,shuide.lyb\}@alibaba-inc.com}
}
  
\begin{document}
\maketitle
\begin{abstract}
Recently, speech-text pre-training methods have shown remarkable success in many speech and natural language processing tasks. However, most previous pre-trained models are usually tailored for one or two specific tasks, but fail to conquer a wide range of speech-text tasks. In addition, existing speech-text pre-training methods fail to explore the contextual information within a dialogue to enrich utterance representations. In this paper, we propose \textbf{S}peech-text dialog \textbf{P}re-training for spoken dialog understanding with \textbf{E}xpli\textbf{C}i\textbf{T} c\textbf{R}oss-Modal \textbf{A}lignment (SPECTRA), which is the first-ever speech-text dialog pre-training model. Concretely, to consider the temporality of speech modality, we design a novel temporal position prediction task to capture the speech-text alignment. This pre-training task aims to predict the start and end time of each textual word in the corresponding speech waveform. In addition, to learn the characteristics of spoken dialogs,  we generalize a response selection task from textual dialog pre-training to speech-text dialog pre-training scenarios. Experimental results on four different downstream speech-text tasks demonstrate the superiority of SPECTRA in learning speech-text alignment and multi-turn dialog context.\footnote{For reproducibility, we release our code and pre-trained model at: \url{https://github.com/AlibabaResearch/DAMO-ConvAI/tree/main/SPECTRA.}}

\end{abstract}

\section{Introduction}

In recent years, speech-text pre-training, which learns universal feature representations from a large training corpus~\citep{chen2018almost,ctal,bapna2021slam}, has achieved significant success in both uni-modal \citep{wav2vec,dosovitskiy2020image} and multi-modal \citep{lu2019vilbert,radford2021learning}  downstream tasks. Existing speech-text pre-training works mainly employed multi-modal self-supervised pre-training objectives, such as cross-modal masked data modeling~\citep{ctal,kang2022self} and cross-modal contrastive learning~\citep{calm,elizalde2022clap}, which align the speech utterance representation to the corresponding text sentence representation.

\begin{figure}[t]
    \centering
    \includegraphics[width=\linewidth]{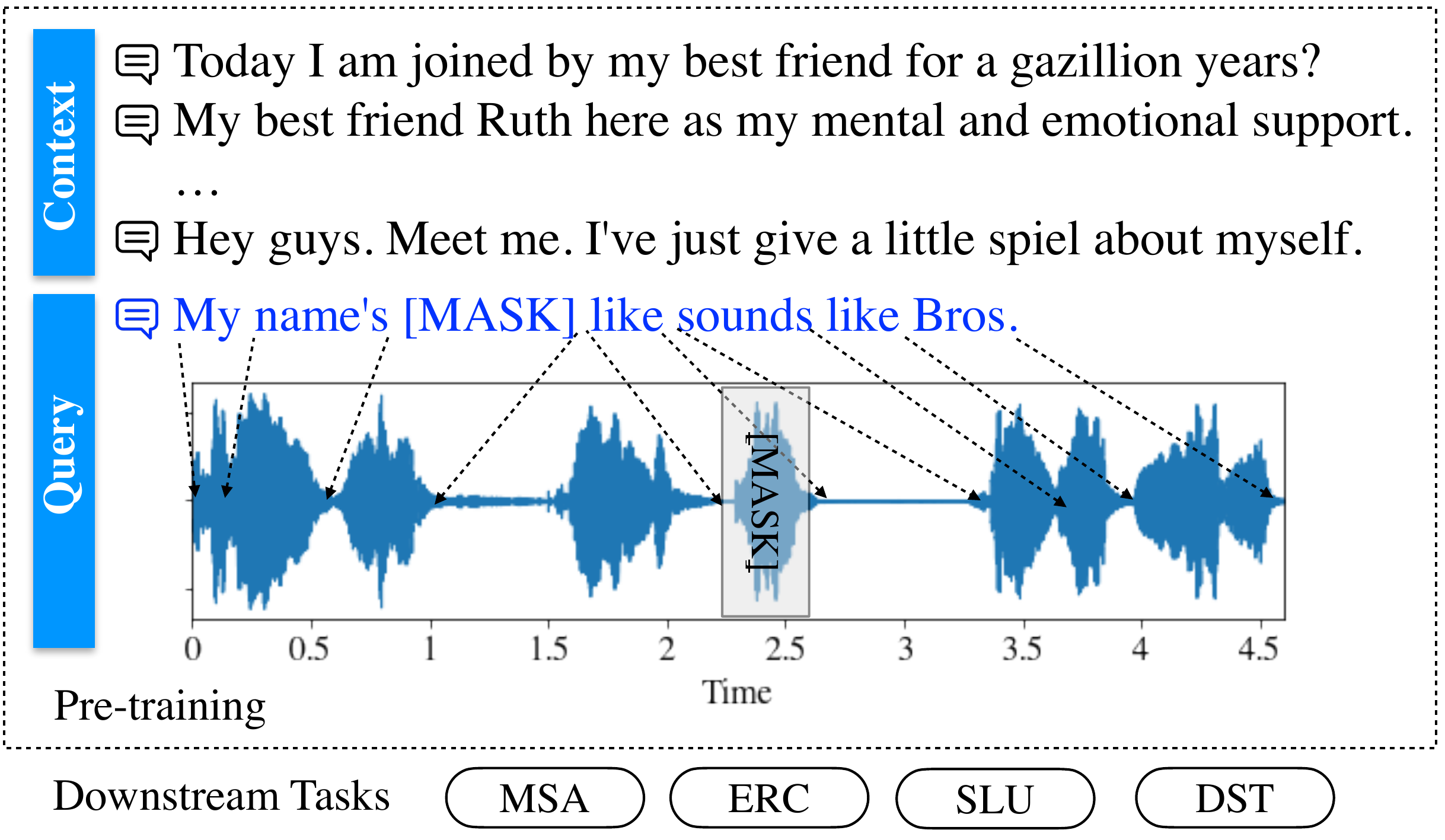}
    \caption{An illustration of SPECTRA, which considers dialogue context and explicit alignment between text and speech during pre-training, and generalizes well on various downstream tasks.}
    \label{fig:running_example}
\end{figure}

Despite the remarkable progress of previous speech-text pre-training models, there are still several technical challenges to constructing an effective and unified speech-text pre-training model for spoken dialog understanding, which are not addressed well in prior works.
First, previous models are mainly tailored for specific speech-text tasks, such as speech-to-text translation \cite{liu2020multilingual} and speech-language understanding \cite{chung2020splat}, failing to conquer a wide range of speech-text tasks. Although \citet{tang2022unified} proposed a unified speech-text pre-training for speech translation and recognition, it fails to exploit the temporality of an input speech sequence and cannot learn the fine-grained speech-text alignment. 

Second, limited exploration has been attempted to bridge the gap between plain speeches/texts and human conversations. In particular, existing speech-text pre-training methods fail to explore the context information within a dialog. Nevertheless, spoken dialog understanding needs to effectively process context information so as to help the system better understand the current utterance, since humans may omit previously mentioned entities/constraints and introduce substitutions to what has already been mentioned.

\begin{figure*}[t]
    \centering
    \includegraphics[width=\linewidth]{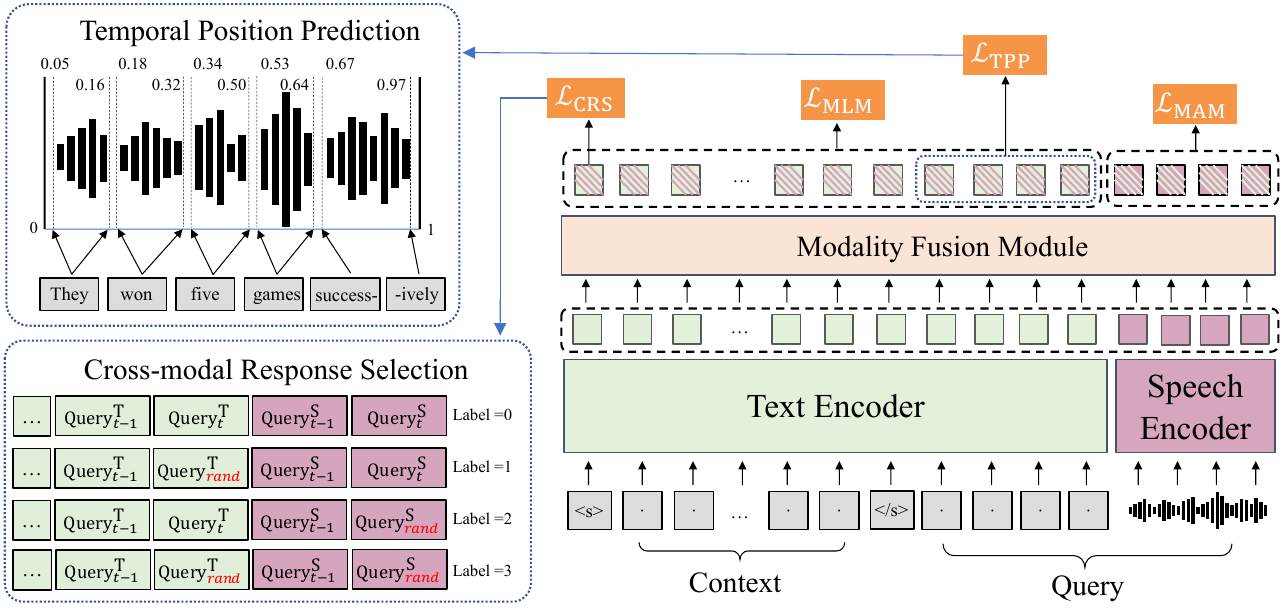}
    \caption{The overview of SPECTRA. The left part shows the illustration of the temporal position prediction task and the cross-modal response selection task. The right part shows the overall structure of the pre-trained model.}
    \label{fig:model}
\end{figure*}

In this paper, we propose \textbf{S}peech-text dialog \textbf{P}re-training for spoken dialog understanding with \textbf{E}xpli\textbf{C}i\textbf{T} c\textbf{R}oss-Modal \textbf{A}lignment (SPECTRA), which is the first-ever speech-text dialog pre-training model. 
We illustrate the framework of our method in Figure \ref{fig:running_example} and details in Figure \ref{fig:model}.
The backbone of SPECTRA is composed of a text encoder, a speech encoder, and a fusion module, learning semantic/acoustic information and the interaction between them, and pre-trained on a large-scale real-world multi-modal (speech-text) dialog corpus. We propose two pre-training objectives to learn better context-aware speech/text representations for spoken dialog understanding \cite{dai2022cgodial, zhang-etal-2022-slot}. Specifically, to consider the temporality of speech modality, we design a novel temporal position prediction task to capture the speech-text alignment by predicting the start and end time of each textual word in the corresponding speech waveform. In addition, to learn the characteristics of spoken dialogs \cite{gao2023unsupervised, qian2023empathetic}, we devise a cross-modal response selection objective to consider the context information within each dialog. 



Our contributions are summarized as follows:
\begin{itemize}
\item To the best of our knowledge, we are the first to propose a speech-text dialog pre-training model for spoken dialog understanding, which fully exploits the characteristics of multi-modal (speech/text) dialogs.  
\item We introduce two pre-training objectives (temporal position prediction and multi-modal response selection) to effectively learn speech-text alignment and dialog context information. 
\item We conduct extensive experiments on five benchmark datasets belonging to four downstream speech-text tasks, including emotion recognition in conversation (ERC), multi-modal sentiment analysis (MSA), spoken language understanding (SLU), and dialog state tracking (DST). We believe that the release of the pre-trained model and source code would push forward the research in this area.
\end{itemize}

\section{Related Work}
\paragraph{Uni-modal Pre-training}
In recent years, pre-trained language models (PLMs), such as BERT \citep{bert}, RoBERTa \citep{roberta}, and GPT \citep{gpt2} have been proposed and applied to many NLP tasks, yielding impressive performances. PLMs benefit from the rich linguistic knowledge in large-scale corpora\cite{he2022galaxy, he2022space2}. Inspired by the success of PLMs in NLP tasks, several speech pre-training models, such as Wav2vec \citep{wav2vec}, HuBERT \citep{hubert}, and WavLM \citep{wavlm}, were proposed to learn high-quality universal speech representations from massive speech data.  


\paragraph{Multimodal Pre-training}
Compared to multi-modal pre-training for vision-and-language tasks, speech-text pre-training is relatively less explored. SpeechBERT \citep{chuang2020speechbert} jointly trained multimodal representations based on a single BERT for spoken question-answering. 
CTAL \citep{ctal} extended the original Transformer to cross-modal by modifying the attention mechanism of the Transformer decoder.
ST-BERT \citep{stbert} combined a pre-trained acoustic model with BERT and took phoneme posterior and subword-level tokenized text as input. 
\citet{self} explored multimodal pre-training model in extremely low-resource data scenarios.
CLAM \citep{calm} employed contrastive and multirate information inherent in audio and lexical inputs to align acoustic and lexical information.
STPT \cite{tang2022unified} proposed a multi-task learning framework to integrate different modalities in speech-text pre-training.

\paragraph{Multimodal Dialog Systems}
The demand for multimodal dialog systems \cite{lin2022duplex} is increasing due to the ubiquitous multimodal data. \citet{liao2018knowledge} presented a knowledge-aware multimodal dialog (KMD) model, which leveraged reinforcement learning to generate human-like responses given multimodal (text-image) dialog context. 
\citet{cui2019user} considered the explicit user requirements in
the attribute level and dynamically encoded the multimodal (text-image) dialog context based on users' attention.
\citet{sunder2022towards} proposed an end-to-end spoken language understanding model, which trained a semantically rich BERT-based conversation model along with a speech-based model.

Different from previous works, SPECTRA is the first-ever speech-text dialog pre-training model, which bridges the gap between plain texts/speeches and human conversations.


\section{Method}

In this section, we introduce the model architecture and pre-training objectives of SPECTRA.

\subsection{The Backbone Architecture}
Figure \ref{fig:model} shows the overall structure of our model SPECTRA, which consists of a text encoder, a speech encoder, and a modality fusion module. During pre-training, we first convert paired text and speech inputs into uni-modal embeddings, which are then fed into the text encoder and speech encoder respectively to obtain uni-modal representations. Finally, we concatenate text representations and speech representations as input of our modality fusion module to get fused representations for speech-text pre-training.

\subsubsection{Data Preparation}
Before diving into our model, we first prepare input text and speech sequences for our model. 
Let $D = \{T_1, T_2, ..., T_n\}$ denotes a conversation with $n$ dialog turns, where every single dialog turn $T_i$ consists of a slice of raw speech waveform $\mathbf{s}_i$ and its corresponding text $\mathbf{t}_i=\{\mathbf{w}_{i1}, \mathbf{w}_{i2}, ..., \mathbf{w}_{im}\}$. 
Here, $\mathbf{w}_{ij}$ is the $j$-th word of $\mathbf{t}_i$, and is annotated with its corresponding start/end time in the speech, denoted as $s_{ij}$/$e_{ij}$. 
$m$ is the sentence length of $\mathbf{t}_i$. 
For each dialog turn $T_i$ where $i > 1$, we construct a sample $\mathbf{X}_i$ with current utterance $T_i=\{\mathbf{t}_i,\mathbf{s}_i\}$, previous $k$ ($k\geq 1$) turns of textual dialog history $\{\mathbf{t}_{i-k},,...,\mathbf{t}_{i-2},\mathbf{t}_{i-1}\}$ and the previous speech dialog history $\mathbf{s}_{i-1}$. 
In this way, each sample $\mathbf{X}_i$ consists of $k$+1 turns of text and 2 turns of speeches, where the speeches correspond to the latest 2 turns of text.
Note that we only use 2 turns of speech in pre-training for efficiency, since the length of speech representation is much longer than its corresponding text representation.

\subsubsection{Text Embeddings}
For each input element, its vector representation is a summation of the corresponding token embedding, absolute position embedding and segment embedding.
Specifically, we first concatenate all text sentences of each sample $\mathbf{X}_i$ in temporal order to construct the text input:
$I_i$=<s>$\mathbf{t}_{i-k}$</s>$\mathbf{t}_{i-k+1}$</s>...</s>$\mathbf{t}_{i-1}$</s>$\mathbf{t}_i$</s>.
Note that we use special token <s> to mark the start of the whole sequence, and </s> to mark the end of each turn.
Then, we encode each token in $I_i$ using a pre-trained RoBERTa \citep{roberta} tokenizer.
We assign learnable segment embedding $\mathbf{e}_{t,1}$ to tokens of $\mathbf{t}_i$ and the last </s> token, and $\mathbf{e}_{t,0}$ for the rest of the tokens. The detailed tokenizing and encoding process is described in Appendix \ref{sec:a2}. We denote $\mathbf{x}_i$ as the input text embeddings of $I_i$.

\subsubsection{Uni-modal Encoders}
\paragraph{Text Encoder} Inspired by the remarkable success of uni-modal pre-trained models on various downstream tasks, we employ RoBERTa \citep{roberta} as our text encoder. We pass $\mathbf{x}_i$ into text encoder to obtain the sequence representations:
\begin{equation}
\mathbf{H}_{t,i}=\text{RoBERTa}(\mathbf{x}_i)
\end{equation}
where $\mathbf{H}_{t,i} \in \mathbb{R}^{n\times d_h}$ denotes the output hidden states of the last layer of RoBERTa, $n$ is the length of input $I_i$, and $d_h$ is the dimension of hidden state.

\paragraph{Speech Encoder} 
We design our speech encoder based on the WavLM structure \citep{wavlm} with three key modules: a feature extractor, a feature projection module and a Transformer encoder module. The feature extractor consists of 8 temporal convolutional layers and a layer normalization. We implemented the first seven convolutional layers to be the same as WavLM, and added another convolutional layer with 512 channels, 5 strides and 5 kernels size, in order to shorten the length of the output speech features. As a result, each output token of speech features represents approximately 200ms of speech with a stride of 100ms. 

The feature projection layer is a layer normalization followed by a fully connected layer converting the size of speech features from 512 to $d_h$. The Transformer encoder module is equipped with a convolution-based relative position embedding layer and 12 WavLM Transformer layers. 
For each sample, we directly input speech waveforms $\mathbf{s}_{i-1}$ and $\mathbf{s}_i$ into our speech encoder, and denote the outputs of the feature projection layer for $\mathbf{s}_{i-1}$ and $\mathbf{s}_i$ as $f_{i-1}$ and $f_i$:
\begin{align}
f_{i-1}&=\text{Proj}(\text{Conv}(\mathbf{s}_{i-1}))\\
f_i&=\text{Proj}(\text{Conv}(\mathbf{s}_i))
\end{align}
Then, we obtain a speech sequence $a_i$ by concatenating $f_{i-1}$ and $f_i$ together with a separation token [SEP] and a starting token [CLS]:
\begin{equation}
a_i=[\text{CLS}] f_{i-1} [\text{SEP}] f_i
\end{equation}
where $a_i\in\mathbb{R}^{(m_{i-1}+m_i+2)\times d_h}$ denotes the concatenated sequence. $m_{i-1}$ and $m_i$ are the lengths of $\mathbf{s}_{i-1}$ and $\mathbf{s}_i$, respectively. We pass $a_i$ as the input of the Transformer encoder module to get the speech sequence representations:
\begin{equation}
\mathbf{H}_{s,i}=\text{WavLM}(a_i)
\end{equation}
where $\mathbf{H}_{s,i}\in\mathbb{R}^{(m_{i-1}+m_i+2)\times d_h}$ denotes the hidden states of the last Transformer layer.

\subsubsection{Modality Fusion Module}
To integrate two modalities, we employ a single self-attention Transformer layer as our modality fusion module. We first concatenate  the text sequence representation $\mathbf{H}_{t,i}$ and the speech sequence representation $\mathbf{H}_{s,i}$ together.
Then, we assign text and speech representations with learnable modality embeddings $\mathbf{e}_{m,0}$ and $\mathbf{e}_{m,1}$ respectively, and add the modality embeddings to the concatenated representations as the input of our modality fusion module. Finally, we obtain output hidden representations of modality fusion module $\mathbf{H}_{i}\in\mathbb{R}^{(n+m_{i-1}+m_i+2)\times d_h}$ as the speech-text joint representations.

\subsection{Pre-training Tasks}
We introduce two novel pre-training objectives for our SPECTRA model, empowering SPECTRA to capture speech-text alignment and multimodal dialog context effectively. 

\subsubsection{Temporal Position Prediction}
Existing speech-text pre-training works mainly learn from prior visual-text pre-training models. 
These works ignore that speeches are temporal sequences, and thus fail to learn fine-grained speech-text alignment. 
In this work, we propose a novel temporal position prediction (TPP) objective, which utilizes the textual part of the hidden representations $\mathbf{H}_{i}$ to predict the starting and ending time of each word in the speech waveform. 

In particular, for each word $\mathbf{w}_{ij}$ in utterance $\mathbf{t}_i$ with its start/end time annotations $s_{ij}$/$e_{ij}$, we denote its first/last token in $\mathbf{H}_{i}$ as $\mathbf{h}_{s_{ij}}$/$\mathbf{h}_{e_{ij}}$. 
The goal of the TPP pre-training objective is to predict its starting and ending time in $\mathbf{s}_i$ with $\mathbf{h}_{s_{ij}}$ and $\mathbf{h}_{e_{ij}}$, respectively. We use squared error loss to optimize the TPP task:
\begin{align}
\mathcal{L}_{\rm TPP}(t_i) &= \frac{1}{2}\left(\left(\mathbf{W}_{start}\mathbf{h}_{s_{ij}} - \frac{s_{ij}}{L_a}\right)^2\right.\\ \notag &+\left.\left(\mathbf{W}_{end}\mathbf{h}_{e_{ij}} - \frac{e_{ij}}{L_a}\right)^2\right)
\end{align}
where $\mathbf{W}_{start},\mathbf{W}_{end}\in\mathbb{R}^{d_h\times 1}$ are learnable parameters. $L_a$ is the maximum speech length limit. By normalizing  $s_{ij}$ and $e_{ij}$ over $L_a$, we guarantee that the starting and ending time falls into [0,1].
Here, we only calculate the TPP loss for the words in the last two turns of dialog (i.e., $\mathbf{t}_{i-1}$ and $\mathbf{t}_i$) for each sample $\mathbf{X}_i$. We calculate the average TPP loss over all words within those two turns as the TPP loss of dialog $\mathbf{X}_i$:
\begin{equation}
\small
\mathcal{L}_{\rm TPP} = \frac{1}{l_{i-1}+l_i}[\sum_j\mathcal{L}_{\rm TPP}(\mathbf{w}_{i-1,j})+\sum_j\mathcal{L}_{\rm TPP}(\mathbf{w}_{i,j})]
\end{equation}
where $l_{i-1}$ and $l_{i}$ denote the total lengths of transcripts $\mathbf{t}_{i-1}$ and $\mathbf{t}_i$ in sample $\mathbf{X}_i$.

\subsubsection{Cross-modal Response Selection}
Inspired by the success of response selection tasks in textual dialog systems \citep{bao2019plato}, we design a cross-modal response selection objective. 
For each sample $\mathbf{X}_i$, we randomly replace the text query $\mathbf{t}_i$ or speech query $\mathbf{s}_i$ with the utterances or speech from other dialogs in the dataset.
In this way, for each sample $\mathbf{X}_i$, we can obtain three kinds of corrupted samples as negatives: (1) only the speech query is randomly substituted; (2) only the text query is randomly substituted; (3) both text and speech queries are randomly substituted. Note that both text and speech queries remain unchanged as positive as illustrated in Figure \ref{fig:model} 

Since the output of the first <s> token can be viewed as the representation of the whole speech-text sample, we apply a softmax function following a fully connected layer on top of the hidden state of token <s> as a four-way classifier, predicting which case the current example belongs to. 
We utilize the cross-entropy loss to optimize the cross-modal response selection task, denoted as $\mathcal{L}_{\rm CRS}$.


\subsubsection{Cross-modal Masked Data Modeling}
Following previous works~\cite{ctal}, we also adopt the cross-modal representations $\mathbf{H}_{f}$ for cross-modal masked language modeling (CMLM) and cross-modal masked acoustic modeling (CMAM) objectives. 
For masked language modeling, we follow the setup of RoBERTa \citep{roberta} to dynamically mask out textual input tokens with a probability of 15\%. For masked acoustic modeling, we follow \citet{baevski2020wav2vec} and \citet{Liu_2020} to mask continuous speech frames. 

We modify the implementation of the original masked acoustic modeling method in previous works to increase the average number of masked speech frames in each sample. 
We provide the details of masked acoustic modeling in Algorithm \ref{algo:mam} in Appendix \ref{sec:a1}. 
The speech token masking step is performed between the feature extractor and feature projection. We employ the cross-entropy loss for the CMLM task ($\mathcal{L}_{\rm CMLM}$) and the mean absolute error loss for the CMAM task ($\mathcal{L}_{\rm CMAM}$).

\subsubsection{Joint Pre-training Objective}
We combine four pre-training objectives to form a joint pre-training objective for speech-text pre-training:
\begin{equation}
\mathcal{L}=\alpha\mathcal{L}_{\rm TPP}+\mathcal{L}_{\rm CRS}+\mathcal{L}_{\rm CMLM}+\mathcal{L}_{\rm CMAM}
\end{equation}

\subsection{Fine-tuning on Downstream Tasks}
We fine-tune SPECTRA on four downstream tasks, including multimodal sentiment analysis (MSA), emotion recognition in conversation (ERC), spoken language understanding (SLU), and dialog state tracking (DST). 

We use the hidden state of <s> token in $\mathbf{H}_i$, denoted as $\mathbf{h}_{i}$ , and pass it through a prediction head with two fully-connected layers and a GELU activation \citep{hendrycks2016gaussian} between them to get the prediction:
\begin{equation}
\mathbf{y}_i=\mathbf{W}^{(2)}\sigma(\mathbf{W}^{(1)}\mathbf{h}_i+\mathbf{b}^{(1)})+\mathbf{b}^{(2)}
\end{equation}
where $\sigma$ denotes the GELU activate function, $\mathbf{W}^{(1)}\in\mathbb{R}^{d_h\times d_h}$, $\mathbf{W}^{(2)}\in\mathbb{R}^{d_h\times d_o}$,
$\mathbf{b}^{(1)}\in\mathbb{R}^{d_h}$,
$\mathbf{b}^{(2)}\in\mathbb{R}^{d_o}$ are new learnable parameters in the fine-tuning stage.
The output size $d_o$ for MSA task is 1, and for ERC and SLU it is the corresponding number of classes.
We adopt the squared error loss as the fine-tuning loss function for MSA. The cross-entropy loss is utilized for the rest of tasks. 

\begin{table*}[]
\centering
\resizebox{0.99\textwidth}{!}{
\begin{tabular}{cllll}
\toprule
\textbf{Task}                                        & \textbf{Dataset} & \textbf{Metric} & \textbf{Previous SOTA} & \textbf{SPECTRA}       \\ \midrule
\multirow{2}{*}{Multimodal Sentiment Analysis (MSA)} & MOSI             & Acc$_2$         & 84.40 (MIB~\citep{mai2022multimodal})            & \textbf{87.50 ({\color{ForestGreen}+3.10})} \\ \cmidrule{2-5} 
                                                     & MOSEI            & Acc$_2$         & 86.20 (BBFN~\citep{han2021bi})          & \textbf{87.34 ({\color{ForestGreen}+1.14})} \\ \midrule
Emotion Recognition in Conversation (ERC)            & IEMOCAP          & Acc             & 66.52 (M2FNET~\citep{chudasama2022m2fnet})         & \textbf{67.94 ({\color{ForestGreen}+1.42})} \\ \midrule
Spoken Language Understanding (SLU)                  & MIntRec          & Acc$_{20}$      & 72.16 (MAG-BERT~\citep{rahman2020integrating})       & \textbf{73.48 ({\color{ForestGreen}+1.32})} \\ \midrule
Dialog State Tracking (DST)                          & SpokenWoz        & JGA             & 20.90 (SPACE+WavLM+TripPy~\citep{si2023spokenwoz})          & \textbf{21.96 ({\color{ForestGreen}+1.06})} \\
\bottomrule
\end{tabular}
}
\caption{The comparison between the key metrics of our model and the previous SOTA method on five datasets. }
\label{tab:main_results}
\end{table*}

\section{Experiments}

\subsection{Pre-training Data}
In this paper, we adopt Spotify100K \citep{clifton-etal-2020-100000} to pre-train SPECTRA, which is a real-world scene speech-text dialog dataset.
Spotify100K contains 105,360 podcast episodes, with nearly 60,000 hours of speeches covering a variety of genres, subject matter, speaking styles, and structure formats.
The corpus also provides automatically-generated word-level textual transcripts, marking the starting and ending time in the speech for each word.

For a fair comparison with previous speech-text pre-training studies, we only use the first 960 hours of speech as well as the corresponding transcripts to pre-train our SPECTRA model.

\subsection{Experimental Setup}
\paragraph{Baselines}
In addition to state-of-the-art downstream models tailored for MSA, ERC, SLU and DST (see Section \ref{sec:msa}-\ref{sec:spokenwoz}), we also compare SPECTRA with three types of pre-training models, including the text modality pre-training model RoBERTa \cite{roberta}, speech modality pre-training model WavLM \cite{wavlm}, and speech-text multimodal pre-training model CTAL \cite{ctal}. 




\paragraph{Experimental Settings during Pre-training}
We use the first 960 hours of speech and textual transcripts of Spotify100K dataset for pre-training. 
We cut the speech waveform into slices of a maximum length of 10 seconds and view each slice with the corresponding transcripts as a single dialog turn, forming 356,380 dialog turns in total. 
By using these dialogs and setting $k$ to a maximum of 7, we construct 350,784 samples, where each sample consists of 2\textasciitilde 8 dialog turns of texts and 2 turns of speeches. 

Besides, we use pre-trained models \textbf{RoBERTa-base} and \textbf{WavLM-base+} to initialize our text and speech encoder, respectively.
Since our speech encoder has one more convolution layer than \textbf{WavLM-base+}, we only initialize the first seven convolution layers with pre-trained parameters and randomly initialize the last layer. Both text and speech encoders have 12 Transformer layers with a hidden size $d_h$ of 768. 
We pre-train our SPECTRA model for 100 epochs on 8 Tesla-A100 GPUs with a batch size of 20 per GPU. We use AdamW \citep{loshchilov2018fixing} to optimize our model with a peak learning rate of $1\times 10^{-4}$ and a linear warmup for the first 1\% of updates. 

\paragraph{Experimental Settings during Fine-tuning}
For SpokenWoz dataset, each dialog turn consist of two utterances, one from the user and the other from the system. For other datasets, each dialog turn is a single utterance. 
For all datasets we truncate the speech length of each dialog turn to a maximum of 10 seconds.
We fine-tune our pre-trained checkpoint on each downstream dataset using an AdamW \citep{loshchilov2018fixing} optimizer with a peak learning rate of $2\times 10^{-5}$ and a cosine annealing warmup.

\subsection{Fine-tuning on MSA}
\label{sec:msa}
For MSA task \cite{hu2022unimse}, our model aims to predict the positive or negative sentiment polarities of the given multi-modal input.
We conduct experiments on two multi-modal datasets MOSI \citep{mosi} and MOSEI \citep{mosei} to evaluate the effectiveness of our model for the MSA task. 
We adopt the accuracy over positive/negative sentiments classification (denoted as Acc$_2$) as the evaluation metric for our model and baselines. The experimental results are reported in Table \ref{tab:main_results}.

From the results, we can observe that our model achieves substantially better performance than previous state-of-the-art (SOTA) methods on both datasets.
In particular, for the MOSI dataset, the accuracy increases by 3.10\% over the strongest baseline MIB~\citep{mai2022multimodal}.
In addition, as shown in Table \ref{tab:abl}, our SPECTRA also significantly outperforms the speech modality pre-training model WavLM and speech-text pre-training model CTAL.  



\begin{table*}[]
\centering
\resizebox{\textwidth}{!}{
\begin{tabular}{ccccccccccc}
\toprule
\multirow{3}{*}{Settings} & \multirow{3}{*}{\begin{tabular}[c]{@{}c@{}}MLM\\ \&\\ MAM\end{tabular}} & \multirow{3}{*}{TPP} & \multirow{3}{*}{CRS} & \multirow{3}{*}{\begin{tabular}[c]{@{}c@{}}Pre-training\\ Data\end{tabular}} & \multirow{3}{*}{\begin{tabular}[c]{@{}c@{}} Turns of \\ Textual \\ Dialog\\ History\end{tabular}} & \multicolumn{2}{c}{MSA}  & ERC   & SLU   & DST   \\ \cmidrule{7-11} 
   &  &   &   &  &    & MOSI  & MOSEI & IEMOCAP   & MIntRec   & SpokenWoz \\ \cmidrule{7-11} 
   &  &   &   &  &    & Acc$_2$   & Acc$_2$   & Acc   & Acc$_{20}$    & JGA \\ 
\midrule
RoBERTa & - & - & - & - & - & 85.67 & 85.88 & 64.53 & 71.24 & 20.76 \\
WavLM   & - & - & - & - & - & 65.85 & 77.90 & 46.90 & 16.63 &  -    \\
CTAL    & $\checkmark$ & - & - & 960h & - & 72.56 & 80.77 & 55.12 & 53.26 & 15.79  \\ \midrule
SPECTRA   & $\checkmark$  & $\checkmark$  & $\checkmark$  & 960h & 7    & \textbf{87.50} & \textbf{87.34} & \textbf{67.94} & \textbf{73.48} & \textbf{21.96} \\
\midrule
(a)     &    &   &   & 0 & 1 & 82.16 & 84.30 & 33.17 & 69.21 & 17.59 \\
(b)    & $\checkmark$  & $\checkmark$  & $\checkmark$  & 360h & 7    & 85.98  & 86.02  & 66.01  & 72.36  &  20.34 \\
(c)    & $\checkmark$  &   & $\checkmark$  & 960h & 7    & 85.52  & 86.19  & 66.15  & 71.69  & 20.87 \\ 
(d)    & $\checkmark$  & $\checkmark$  &   & 960h & 7    & 87.35  & 86.85  & 65.94  & 72.58  & 20.45  \\ 
(e)    & $\checkmark$  & $\checkmark$  & $\checkmark$  & 960h & 1 & 87.20   & 86.93  & 64.98  & 73.03 & 19.78 \\ 
 \bottomrule
\end{tabular}
}
\caption{Ablation test results. Here, setting (a) and (b) mean w/o multi-modal pre-training and using less pre-training data. Setting (c), (d) and (e) indicate w/o TTP task, w/o CRS task and w/o full turns of textual history, respectively.}
\label{tab:abl}
\end{table*}

\subsection{Fine-tuning on ERC}

ERC task requires the model to predict the emotion category of an utterance given a speech clip with its transcripts and dialog history. 
Here, we fine-tune our model with the widely-used IEMOCAP dataset \citep{iemocap}, and follow the settings with \citet{chudasama2022m2fnet} to perform a 6-way classification task. 
For each sample, we construct 11 turns of text and 2 turns of speech with a maximum text length of 512.

In Table \ref{tab:main_results}, we report the accuracy of six-way classification for our model and previous SOTA method M2FNET \citep{chudasama2022m2fnet}.
In addition, from Table \ref{tab:abl}, we can observe that our method outperforms uni-modal pre-training models, as well as speech-text pre-training baseline CTAL. 
Compared with the uni-modal baselines RoBERTa and WavLM, our model benefits from multi-modal pre-training tasks that capture interactions and alignments between modalities.
Compared with CTAL, our model is equipped with better speech-text alignment and multi-turn dialog context information with the help of TPP and CRS pre-training tasks. 




\subsection{Fine-tuning on SLU}
We also conduct experiments on the spoken language understanding (SLU) task, 
which aims to predict the user intent \cite{lin2019deep} given a spoken utterance with the textual transcript. 
We use MIntRec \citep{mintrec} as the experimental dataset for SLU and adopt classification accuracy for the evaluation metric. 

From Table \ref{tab:main_results} and \ref{tab:abl}, we can observe that SPECTRA obtains significantly better results than previous methods. In particular, our SPECTRA model improves the results of RoBERTa and the previous SOTA method MAG-BERT \citep{rahman2020integrating} by 1.55\% and 2.47\%, respectively. 
Compared to WavLM and CTAL, our model can capture semantic information in textual data and the context information within each dialog. 

\subsection{Fine-tuning on DST}
\label{sec:spokenwoz}

For dialogue state tracking, we use a large-scale, cross-modal dataset called SpokenWoz \citep{si2023spokenwoz}. The dataset was collected by crowdsourcing recordings through phone calls using the Appen platform\footnote{https://appen.com/}. Transcriptions were obtained using a commercial ASR system, and speech-text pairs were annotated using a schema similar to MultiWoz \citep{multiwoz}.  SpokenWoz consists of 204k turns, 5.7k dialog, and 249 hours of recordings.
We adopt joint goal accuracy (JGA) as the evaluation metric, which compares the predicted and ground-truth dialogue states at each turn. 
We follow Trippy \citep{trippy} and substitute its context model BERT with our SPECTRA model. 

As shown in Table \ref{tab:main_results}, our model outperforms the previous SOTA method, SPACE+WavLM+TripPy. 
In addition, our model also surpasses the three pre-training baselines by a noticeable margin. 
This demonstrates better speech-text alignment is critical to tackling complicated conversations.

\begin{figure*}[t]
    \centering
    \subfigure[Case \#1: SPECTRA]{
        \centering
        \includegraphics[width=0.52\textwidth]{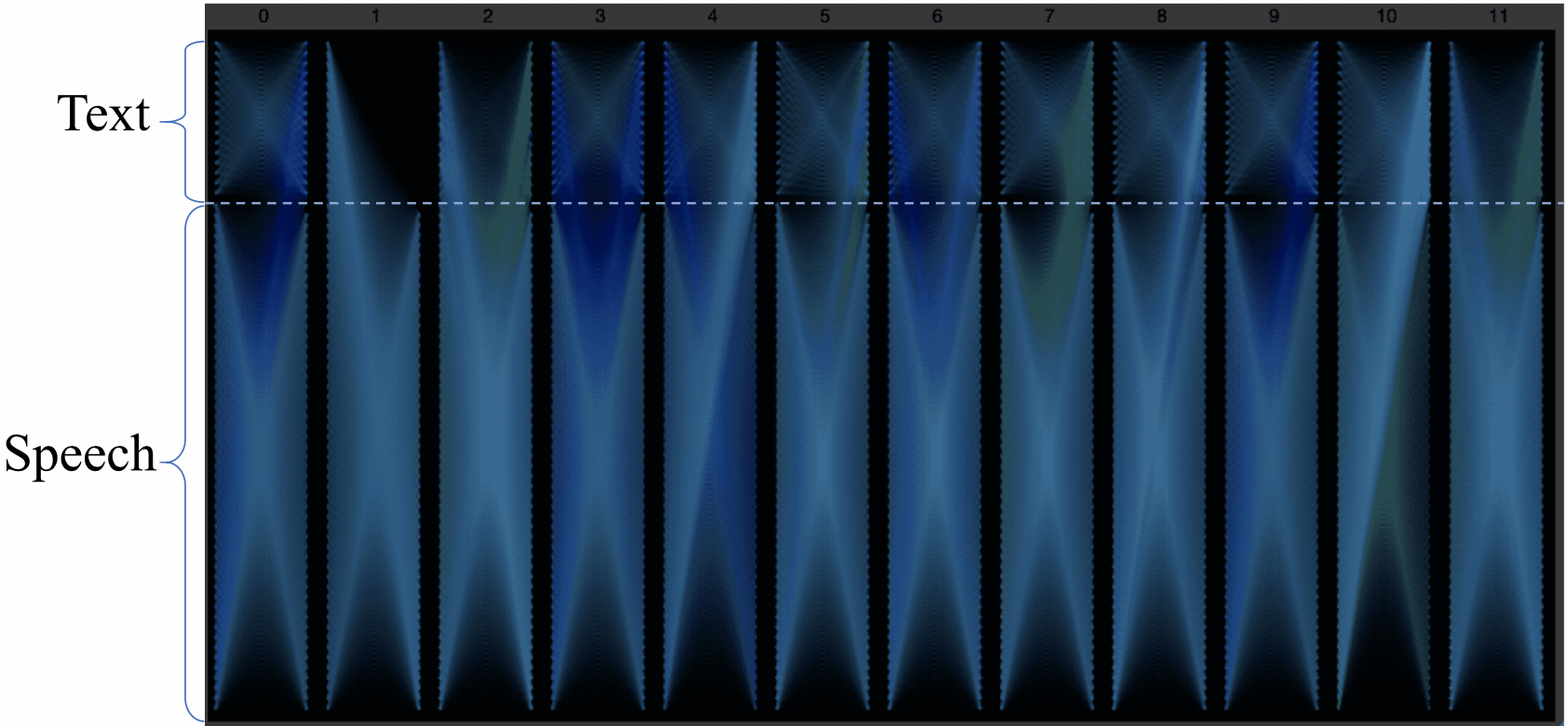}
        \label{sf:m74}
    }
    \subfigure[Case \#1: w/o TPP]{
        \centering
        \includegraphics[width=0.45\textwidth]{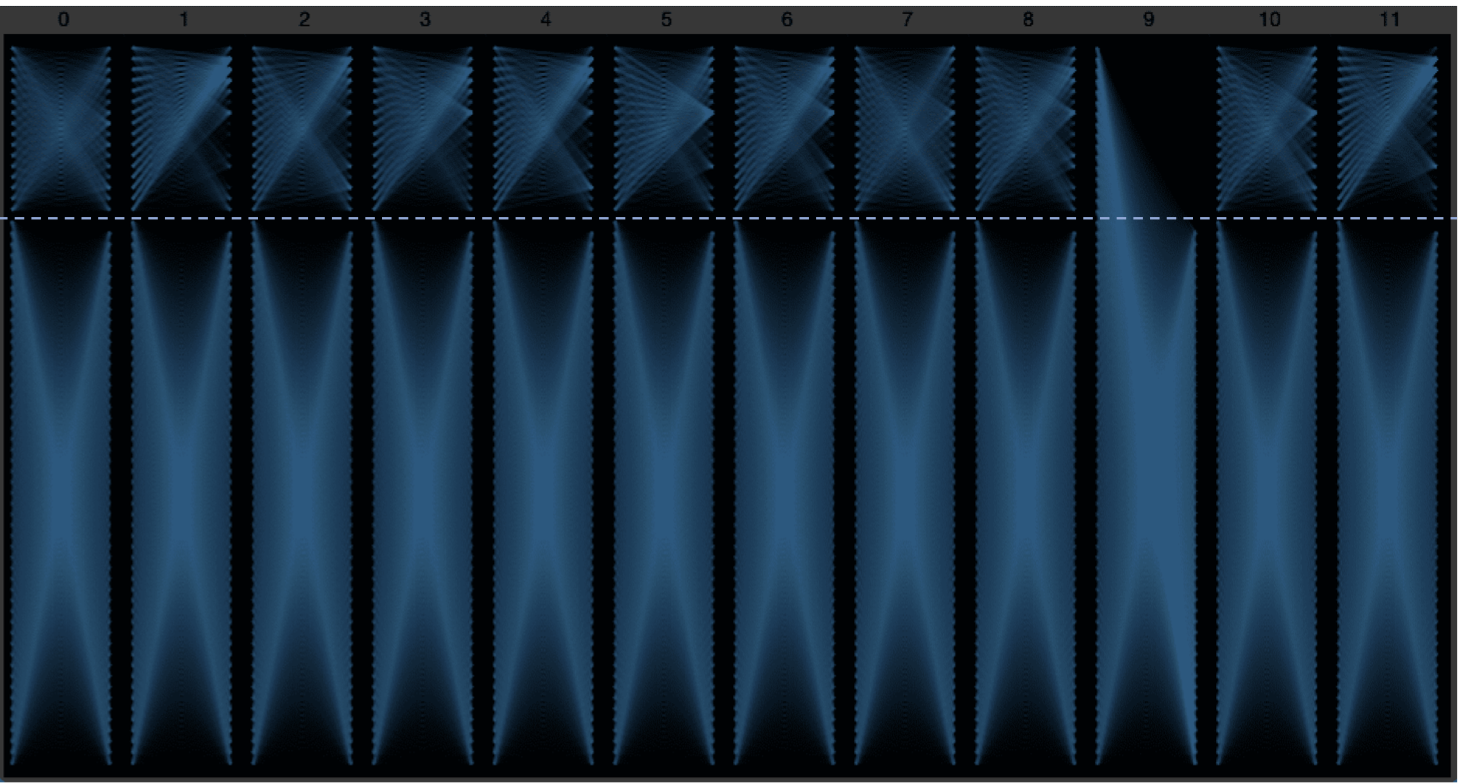}
        \label{sf:m74b}
    }

    \subfigure[Case \#2: SPECTRA]{
        \centering
        \includegraphics[width=0.52\textwidth]{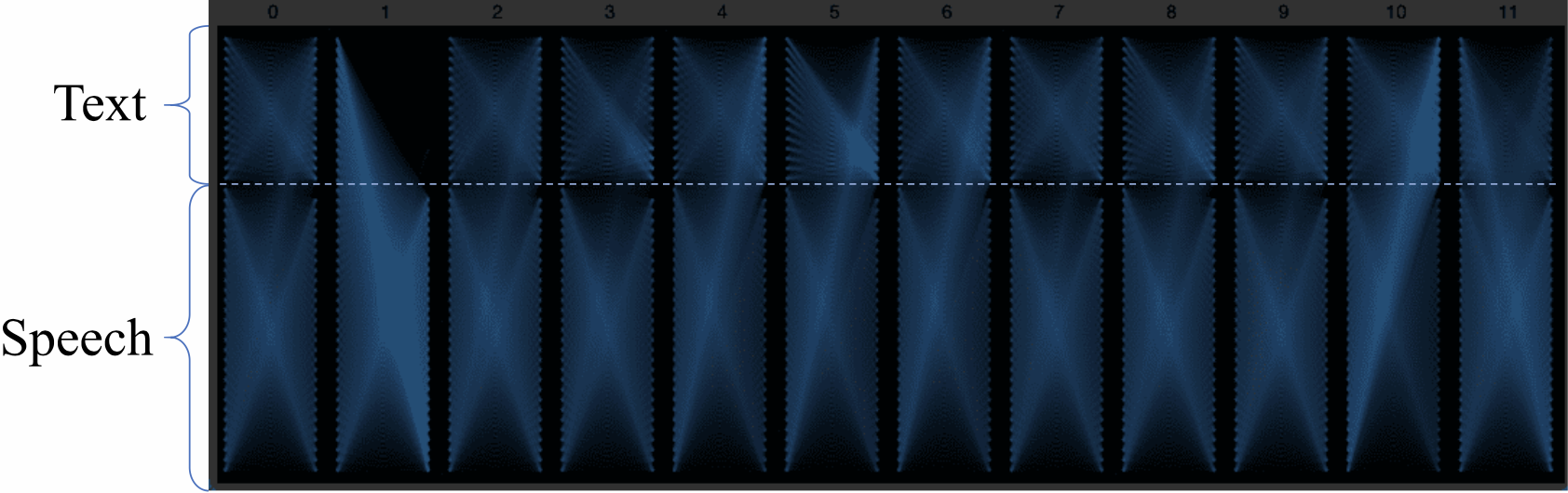}
        \label{sf:m40}
    }
    \subfigure[Case \#2: w/o TPP]{
        \centering
        \includegraphics[width=0.45\textwidth]{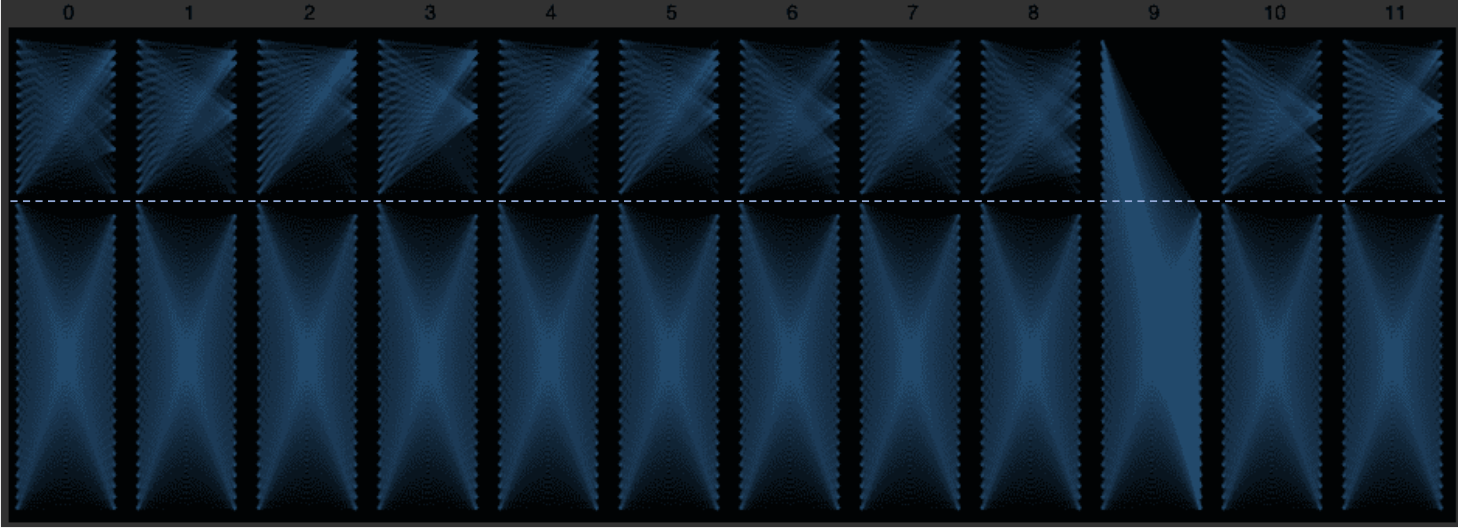}
        \label{sf:m40b}
    }
    \caption{Visualization of self-attention weights of the fusion module in our model and the model pre-trained without TPP (w/o TPP). The upper and lower tokens stand for text and speech tokens, respectively.}
    \label{fig:case}
\end{figure*}

\begin{table*}[t]
\resizebox{\textwidth}{!}{
\begin{tabular}{c|c|c|c|c}
\toprule
\textbf{Case} & \textbf{Ground-Truth} & \textbf{By SPECTRA} & \textbf{By “w/o TPP”} & \textbf{Given Text}                                                \\ \midrule
\#1 & Leave           & {\color{ForestGreen}\checkmark} Leave               & {\color{red}\xmark} Complain              & I am going to sit in traffic for 45 minutes and return this. \\ \midrule
\#2 & Inform          & {\color{ForestGreen}\checkmark} Inform              & {\color{red}\xmark} Praise                & That's Tommy. He's lead organizer, total badass.             \\ 
\bottomrule
\end{tabular}
}
\caption{Intent prediction results on test samples from the MIntRec dataset.}
\label{tab:case}
\end{table*}

\section{Analysis}

\subsection{Ablation Study}

To better understand the effectiveness of our SPECTRA pre-training method, we investigate the influence of pre-training components and dialog history on the overall performance of SPECTRA.
We report the ablation test results in Table \ref{tab:abl}.

\paragraph{Impact of Pre-training}
To demonstrate the efficiency of multi-modal pre-training, we directly use uni-modal encoders and randomly initialize the modality fusion module. 
We observe a significant performance drop by comparing (a) ``w/o multi-modal pre-training'' to other pre-training settings on all five datasets. 
In particular, setting (a) directly collapses on the ERC task, which is a complicated and conversational scenario. 
This verifies the necessity of cross-modal pre-training and aligning speech-text modalities. 
In addition, by comparing SPECTRA and setting (b) ``using less pre-training data'', we can find that using more pre-training data can further improve the performance of our model.

\paragraph{Impact of TPP and CRS}
By comparing the setting (c) ``w/o TPP'' to SPECTRA, the performances on all five datasets drop to different extents, which verifies the generalization and  effectiveness of our TPP pre-training task. Specifically, the performance drops significantly on SpokenWoz, which requires the model to have a stronger ability to align two modalities. This demonstrates that our TPP pre-training task empowers the model with stronger alignment modeling ability.
For setting (d) ``w/o CRS'' with SPECTRA, the performance drops significantly on multi-turn dialog tasks such as ERC and DST. This suggests that the CRS task is essential to model multi-turn dialog context.

\paragraph{Impact of Dialog History}
In setting (e) ``using 1 turn of textual dialog history'', each instance consists of 2 turns of paired speech and text.
The model performance drops substantially on ERC and DST downstream tasks by comparing it with SPECTRA. This demonstrates that increasing dialog history in the pre-training stage is beneficial to the tasks that require multi-turn dialog context.

\subsection{Case Study}
To have a straightforward understanding of how we learn cross-modal interaction in our proposed SPECTRA model, we conduct a case study by providing two cases sampled from the MIntRec dataset. These two cases are incorrectly predicted by the model pre-trained without TPP but correctly predicted by our SPECTRA model.
In Figure \ref{fig:case}, we visualize the self-attention weights of the fusion layer in our model as well as the model pre-trained without TPP (denoted as w/o TPP). From Figure \ref{sf:m74} and \ref{sf:m40}, we observe that there are rich cross-modal interactions in the fusion layer of the proposed SPECTRA model. Our model can capture fine-grained information between text and speech for more accurate classification.
In contrast, we also visualize the self-attention weights of the w/o TPP model in Figure \ref{sf:m74b} and \ref{sf:m40b}. Both cases show that text and speech sequences seldom connect to each other in self-attention layers. 

In Table \ref{tab:case}, we also illustrate the intent prediction results obtained by SPECTRA and w/o TPP. From the results, we can observe that our model can attend to both text and speech sequences effectively to predict correct intent results. However, w/o TPP is confused by the wrong labels since it hardly attends to speech tokens, which indicates that it has the propensity to omit useful information that exists in speech exclusively.




\section{Conclusion}
In this paper, we proposed our model SPECTRA, the first speech-text dialog pre-training model. Considering the temporality of speech and text modalities, we introduced a novel temporal position prediction pre-training task to learn word-level speech-text alignment. To capture multi-modal dialog context in our model, we generalized the response selection task into multi-modal scenarios. Extensive experiments show that our pre-training method can learn better cross-modal interactions as well as multi-modal contextual information and significantly outperformed other strong baselines. In the future, 
we would like to extend speech-text dialog pre-training to more modalities or generative tasks.

\section*{Limitations}

We analyze the limitations of this work, so as to
further improve the performance of our model in
future work. Based on our empirical observation, we
reveal several limitations, which can be divided
into two primary categories. (1) First, our proposed SPECTRA method relies on large-scale spoken dialog corpora with explicit word-level speech-text alignment annotation, such as Spotify100K. This limits the generality of our model on more spoken dialog corpora. 

In the future, we would like to develop a semi-supervised pre-training method to leverage both labelled and unlabeled datasets. (2) Second, our method is mainly designed for speech-text understanding and has not been fully explored for generative tasks. We plan to devise dialog generation per-training objective to empower the model with better generation ability. (3) Third, the work only involves speech and text modalities. We are interested in handling more modalities, such as images or videos, to enrich cross-modal information in joint representations.



\section*{Acknowledgements}
Min Yang was partially supported by National Key Research and Development Program of China (2022YFF0902100), Shenzhen Science and Technology Innovation Program (KQTD20190929172835662), Shenzhen Basic Research Foundation (JCYJ20210324115614039 and JCYJ20200109113441941), and NSFC (no. 92270122). This work was supported by Alibaba Group through Alibaba Innovative Research Program.

\bibliography{custom}
\bibliographystyle{acl_natbib}
\appendix

\begin{algorithm*}
    \caption{Method to mask speech tokens}
    \begin{algorithmic}[1]
	\Require the output of temporal convolution layer $\mathbf{f}=\text{Conv}(\mathbf{s})$ with the shape of $l\times 512$, where $\mathbf{s}$ is the input speech waveform.
        \Ensure Masked convolutional feature output $\Tilde{\mathbf{f}}$ and masked indices $m_{\mathbf{f}}$.
        \State $\Tilde{\mathbf{f}}=\mathbf{s}$; $m_{\mathbf{f}}=[0,0,...,0]$ with the length of $l$.
        \State Randomly picks an integer $n$ from [20, 50] as the length of masked continuous speech frames.
	\For{$i=0$; $i<l$; $i++$}
            \State Draw a random number $r$ from $U(0,1)$;
            \If{$r<0.15$}
                \Comment{Mask the continuous speech frames from index $i$ to $i+n-1$}
                \State $m_{\mathbf{f}}[i:i+n]=1$;
                \For{$j=0$; $j<n$ \textbf{and} $i+j<l$; $j++$}
                    \State Draw a random number $t$ from $U(0,1)$;
                    \If{$t<0.8$}
                        \State $\Tilde{\mathbf{f}}[i+j]=\mathbf{0}$;
                    \ElsIf{$t<0.9$}
                        \State Replace $\Tilde{\mathbf{f}}[i+j]$ with a random speech frame in $\mathbf{f}$;
                    \EndIf
                \EndFor
                \State $i=i+n-1$
            \EndIf
        \EndFor
    \end{algorithmic}
    \label{algo:mam}
\end{algorithm*}

\section{Implementation details of the tokenizer}
\label{sec:a2}
We describe how we convert each $I_i$ into input embeddings $\mathbf{x}_i$. First, we split the sequence $I_i$ into list of tokens using a BBPE algorithm \cite{radford2019language} and convert each token into its index according to the dictionary of our tokenizer. Then, we pass the token indexes to the pre-trained token embedding layer of RoBERTa model to get the token embedding of each token. Finally, we sum up the token embedding, the absolute positional embedding and the segment embedding ($\mathbf{e}_{t,0}$ or $\mathbf{e}_{t,1}$) to get the input text embedding of every token in $I_i$.

\section{Method to Mask Speech Tokens}
\label{sec:a1}

We report our method to mask speech tokens in Algorithm \ref{algo:mam}. We note that masked speech tokens are set to 0 at 80\% of the time, a random token 10\%, and an un-altered 10\%. In our experiments, the maximum length of speech features $f_i$ is 99 since the longest slice of our speech input is 10 seconds. We estimate the expectation of the number of masked frames of our method and the original MAM method proposed by \citet{Liu_2020} by simulating both masking steps for 1,000,000 times and calculating the average number of masked tokens. Simulation results show that our method masks approximately 57\% of tokens in the sequence, while the original MAM method masks around 15\%.


\end{document}